**Original Paper**

# Quality of Answers of Generative Large Language Models vs Peer Patients for Interpreting Lab Test Results for Lay Patients: Evaluation Study

Zhe He[1,*], PhD, MSc; Balu Bhasuran[1], PhD; Qiao Jin[2], MD; Shubo Tian[2], PhD; Karim Hanna[3], MD; Cindy Shavor[3], MD; Lisbeth Garcia Arguello[3], MD; Patrick Murray, MD[3]; Zhiyong Lu, PhD[2]

[1]School of Information, Florida State University, Tallahassee, Florida, USA
[2]National Center for Biotechnology Information (NCBI), National Library of Medicine (NLM), National Institutes of Health, Bethesda, Maryland, USA
[3]Morsani College of Medicine, University of South Florida, Tampa, Florida, USA

[*]Corresponding author:

Zhe He, PhD
School of Information
College of Communication and Information
Florida State University
Tallahassee, Florida, USA 32306-2100
Tel: 850-644-5775
Email: zhe@fsu.edu

## Abstract

**Background:** Even though patients have easy access to their electronic health records and lab test results data through patient portals, lab results are often confusing and hard to understand. Many patients turn to online forums or question and answering (Q&A) sites to seek advice from their peers. However, the quality of answers from social Q&A on health-related questions varies significantly, and not all the responses are accurate or reliable. Large language models (LLMs) such as ChatGPT have opened a promising avenue for patients to get their questions answered.

**Objective:** We aim to assess the feasibility of using LLMs to generate relevant, accurate, helpful, and unharmful responses to lab test-related questions asked by patients and to identify potential issues that can be mitigated with augmentation approaches.

**Methods:** We first collected lab test results related question and answer data from Yahoo! Answers and selected 53 Q&A pairs for this study. Using the LangChain framework and ChatGPT web portal, we generated responses to the 53 questions from four LLMs including GPT-4, Meta LLaMA 2, MedAlpaca, and ORCA_mini. We first assessed the similarity of their answers using standard QA similarity-based evaluation metrics including ROUGE, BLEU, METEOR, BERTScore. We also utilized an LLM-based evaluator to judge whether a target model has higher quality in terms of relevance, correctness, helpfulness, and safety than the baseline model. Finally, we performed a manual evaluation with medical experts for all the responses of seven selected questions on the same four aspects.

**Results:** Regarding the similarity of the responses from 4 LLMs, where GPT-4 output was used as the reference answer, the responses from LLaMa 2 are the most similar ones, followed by LLaMa 2, ORCA_mini, and MedAlpaca. Human answers from Yahoo data were scored lowest and thus least similar to GPT-4-generated answers. The results of Win Rate and medical expert evaluation both showed that GPT-4's responses achieved better scores than all the other LLM responses and human responses on all the four aspects (relevance, correctness, helpfulness, and safety). However, LLM responses occasionally also suffer from lack of interpretation in one's medical context, incorrect statements, and lack of references.

**Conclusions:** By evaluating LLMs in generating responses to patients' lab test results related questions, we find that compared to other three LLMs and human answer from the Q&A website, GPT-4's responses are more accurate, helpful, relevant, and safer. However, there are cases that GPT-4 responses are inaccurate and not individualized. We identified a number of ways to improve the quality of LLM responses including prompt engineering, prompt augmentation, retrieval augmented generation, and response evaluation.

# Introduction

## Background

In 2021, the US spent $4.3 trillion on healthcare, and 53% of which is attributed to unnecessary use of hospital and clinic services [1,2]. Ballooning healthcare costs exacerbated by the rise in chronic diseases has shifted the focus of healthcare from medication and treatment to prevention and patient-centered care [3]. In 2014, the US Department of Health and Human Services [4] mandated that patients be given direct access to their lab test results. This improves the ability of patients to monitor results over time, follow up on abnormal test findings with their providers in a more timely manner, and prepare them for follow up visits with their doctors [5]. To help facilitate shared decision-making, it is critical for patients to understand the nature of their lab test results within their medical context in order to have meaningful encounters with healthcare providers. With shared decision-making, clinicians and patients can work together to devise a care plan that balances clinical evidence on risks and expected outcomes with patient preferences and values. Current workflows in electronic health records (EHRs) with the 21st Century Cures Act [6] allow patients to have direct access to notes and lab results. In fact, accessing lab results is the most frequent activity patients do when they use patient portals [5],[7]. However, despite the potential benefits of patient portals, merely providing patients access to their records is insufficient for improving patient engagement in their care because lab results can be highly confusing and access may often be without adequate guidance or interpretation [8]. Lab results are often presented in tabular format, similar to the format seen by clinicians [9,10]. The way lab results are presented (e.g., not distinguishing between excellent and close-to-abnormal values) may fail to provide sufficient information about troubling results or fail to prompt patients to seek medical advice from their doctors. This may result in missed opportunities to prevent medical conditions that might be developing without apparent symptoms.

Various studies have found a significant inverse relationship between health literacy and numeracy and the ability to make sense of lab results [11–14]. Patients with limited health literacy are more likely to misinterpret or misunderstand their lab results (either overestimate or underestimate their results), which in turn, may delay them in seeking critical medical attention [5,7,13,14]. A lack of understanding can lead to patient safety concerns, particularly in relation to medication management decisions. Giardina et al. [15] conducted interviews with 93 patients and found that nearly two-thirds did not receive any explanation of their lab results, and 46% conducted online searches to understand their results better. Another study found that patients who were unable to assess the gravity of their test results were more likely to seek information on the Internet or just wait for their doctor to call [14]. There are also potential results where a lack of urgent action can lead to poor outcomes. For example, lipid panel is a commonly ordered lab test that measures the amount of cholesterol and other fats in the blood. If left untreated, high cholesterol can lead to heart disease, stroke, coronary heart disease, sudden cardiac arrest, peripheral artery disease, and microvascular disease [16,17]. When patients have difficulty in understanding lab test results from patient portals but do not have readily access to medical professionals, they often turn to online sources to answer their questions. Among different online sources, social Q&A websites allow patients to ask for personalized advice in an elaborative way or pose questions for real humans. However, the quality of answers on health-related questions on social Q&A varies significantly, and not all responses are accurate or reliable [18,19].

Previous studies, including our own, have explored different strategies for presenting numerical data to patients – e.g., using reference ranges, tables, charts, color, text, and numerical data with verbal explanations, etc. [9,12,20,21]. Researchers have also studied ways to improve patients' understanding of lab results. Kopanitsa [22] studied how patients perceive interpretations of lab results automatically generated by a clinical decision support system. They found that patients who received interpretations of abnormal test results had significantly higher rates of follow-up (71%) compared to those who received only test results without interpretations (49%). Patients appreciate the timeliness of automatically generated interpretations compared to interpretations they can receive from a doctor. Zikmund-Fisher et al. [23] surveyed 1,618 adults in the US to assess how different visual presentations of lab results influence their perceived urgency. They found that a visual line display, which included both the standard range and a harm anchor reference point that many doctors may not consider as particularly concerning, reduced the perceived urgency of close-to-normal alanine aminotransferase and creatinine results (p-value < .001). Morrow et al. [24] investigated whether providing

verbally-, graphically-, and video-enhanced context for patient portal messages about lab results can improve responses to the messages. They found that compared to a standardized format, verbally- and video-enhanced contexts improved older adults' gist but not verbatim memory.

Recent advances in AI-based large language models (LLMs) have opened new avenues for enhancing patient education. LLMs are advanced AI systems that use deep learning techniques to process and generate natural language (e.g., ChatGPT and GPT-4 developed by OpenAI) [25]. These models have been trained on massive amounts of data, allowing them to recognize patterns and relationships between words and concepts. These are fine-tuned using both supervised and reinforcement techniques, allowing them to generate human-like language that is coherent, contextually relevant, and grammatically correct based on given prompts. While LLMs such as ChatGPT have gained popularity, a recent study by the European Federation of Clinical Chemistry and Laboratory Medicine Working Group on AI showed that these may give superficial or even incorrect answers to lab test results related questions asked by professionals and thus, cannot be used for diagnosis [26]. Another recent study by Munoz-Zuluaga et al. [27] evaluated the ability of GPT-4 in answering lab results interpretation questions from physicians in the laboratory medicine field. They found that among 30 questions about lab results interpretation, GPT-4 answered 46.7% of them correctly, provided incomplete/partially correct answers to 23.3% questions, and answered 30% of them incorrectly or irrelevantly. In addition, they found that ChatGPT's response is not sufficiently tailored to the case or clinical questions represented to be useful for clinical consultation.

According to our prior analysis of lab test questions on a social Q&A website [28,29], when patients ask lab test results related questions online, they often focus on specific values, terminologies, or the cause of abnormal results. Some of them may provide symptoms, medications, medical history, and life style information along with lab test results. Previous studies only evaluated ChatGPT's responses of lab test questions from physicians [26], [27], or its ability in answering Yes/No questions [30]. To the best of our knowledge, there is no prior work that evaluates the ability of LLMs in answering lab test questions raised by patients in social Q&A. Hence our goal is to compare the quality of answers from LLMs and social Q&A users to lab test related questions and explore the feasibility of using LLMs to generate relevant, accurate, helpful, and unharmful responses to patients' questions. In addition, we aim to identify potential issues that can be mitigated with augmentation approaches.

## Methods

### Overview

Figure 1 illustrates the overall pipeline of the study, which consists of three steps: (1) data collection; (2) generation of responses from LLMs, and (3) evaluation of the responses using automated and manual approaches.

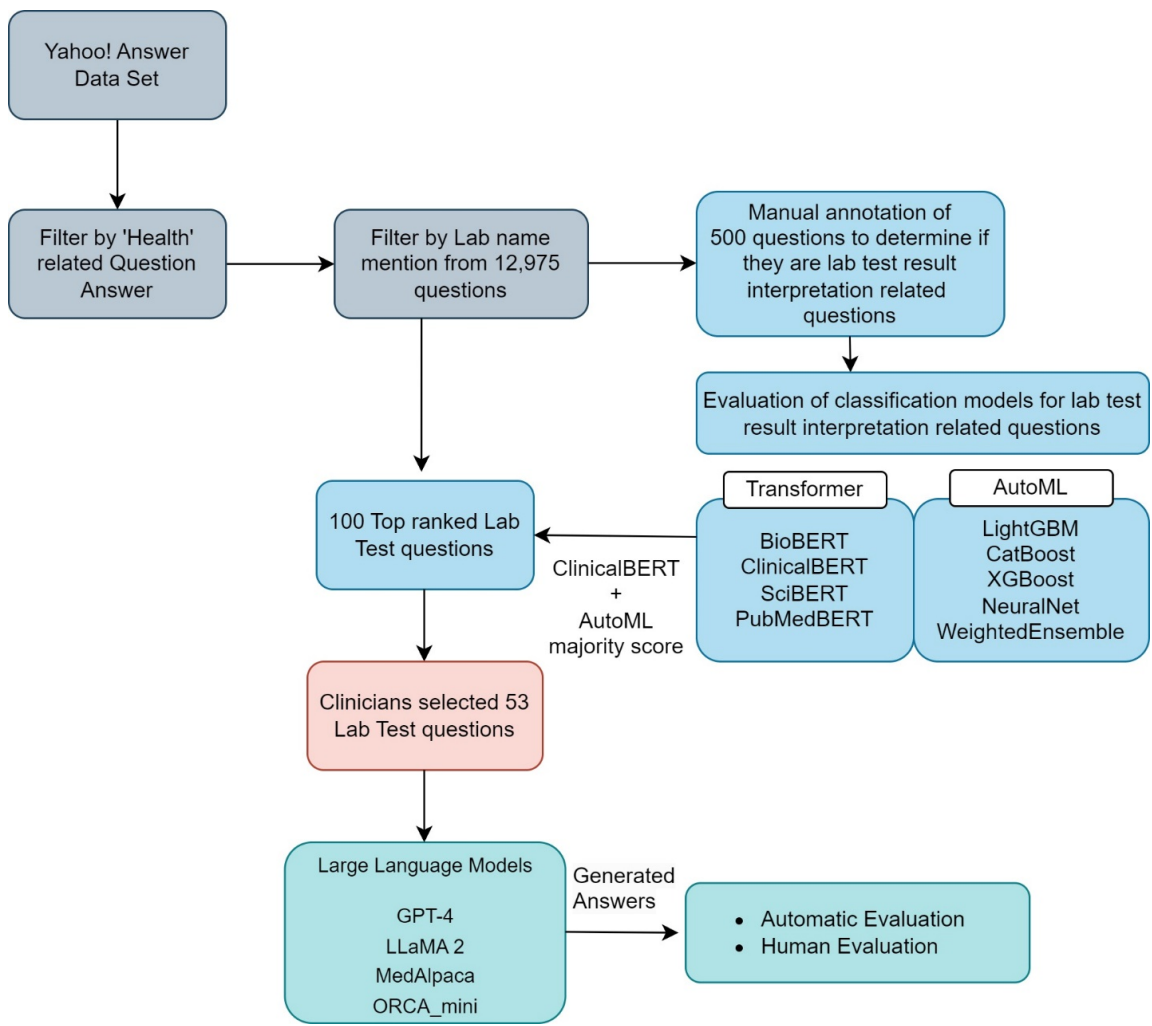

**Figure 1.** Schematic representation of the study pipeline

**Data Collection**

Yahoo! Answer is a community Q&A forum. Its data include questions, responses, and rating of the responses by other users. A question may have more than one answer. We used the answer with the highest rating as the chosen answer. To prepare the dataset for this study, we first identified 12,975 questions that contain one or more lab test names. In our previous work [31], we have annotated key information about lab test results using 251 articles from a credible health information source AHealthyMe.com. The key information included lab test names, alternative names, normal value range, abnormal value range, conditions of normal ranges, indications, actions. However, questions that mention a lab test name may not be about interpretation of test results. To identify questions that are about lab test results interpretation, three undergraduate students in the pre-med track were recruited to manually label 500 randomly chosen questions whether the questions are about lab results interpretation or not. We then trained four transformer-based classifiers (BioBERT [32], ClinicalBERT [33], SciBERT [34] and PubMedBERT [35]) and various automated machine learning models (XGBoost, NeuralNet, CatBoost, WeightedEnsemble, and LightGBM) to automatically identify lab result interpretation related questions from all the 12,975 questions. Then, we worked with primary care physicians to select 53 questions that contain results of blood or urine lab tests on major panels including complete blood count (CBC), metabolic panel, thyroid function, early menopause, and lipid panel.

**Generating Responses from LLMs**

We identified four generative LLMs including OpenAI ChatGPT (GPT-4 version) [36], Meta LLaMA 2 [37], MedAlpaca [38], and ORCA_mini [39] to evaluate in this study.

GPT-4 [36] is the fourth generation Generative Pre-trained Transformer (GPT) model from OpenAI. GPT-4 is a large-scale, multimodal large language model developed using reinforcement learning feedback from both human and AI. The model is reported with human like accuracy in various downstream tasks such as question answering, summarization, and other information extraction tasks based on both text and image data.

LLaMA 2 [37] (Large Language Model Meta AI) is the second generation open-source LLM from Meta, pre-trained using 2 trillion tokens with 4096 token length. Meta released three versions of LLaMa 2 versions such 7B, 13B and 70B parameters with a fine-tuned models LLaMA 2 Chat. The LLaMA 2 models reported with high accuracy on many benchmarks including MMLU, Programming code interpretation, Reading Comprehension, and Open Book QA compared to other open sourced LLMs.

MedAlpaca [38] is an open-source LLM, developed by expanding existing LLM models Stanford Alpaca and AlpacaLoRA by fine tuning it on variety of medical texts. The model is developed as a medical chat bot within the scope of question answering and dialogue application using various medical resources such as medical flash cards, Wikidoc Patient Information, Stack exchange health, USMLE, MEDIQA, PubMed Health Advice and ChatDoctor etc.

ORCA_mini [39] is an open sourced LLM, trained using data and instruction from various open sourced LLMs such as WizardLM (70K), Alpaca (52K) and Dolly-V2 (15K). ORCA_mini is a fine tuned model from OpenLLaMA-3B which is Meta AI's LLaMA 7B trained on the RedPajama dataset. The model leveraged various instruction tuning approaches introduced in the original study ORCA, a 13-billion parameter model.

LangChain [40] is a framework for developing applications by leveraging large language models. LangChain allows users to connect to a language model from repository such as Hugging Face, deploy that model locally and interact with it without any restriction. LangChain enables the user to perform downstream tasks such as Question Answering over specific documents, deploy Chatbots and Agents using the connected LLM. With the rise of open sourced LLMs, LangChain is emerging as a robust framework to connect with various LLMs for user specific tasks.

We used Hugging Face repository of the three LLMs Meta LLaMA 2 [37], MedAlpaca [38] and OCRA_mini [39] to download the model weights and used LangChain input prompts to the models to generate the answers for all the 53 selected questions. The answers were generated in a zero-shot setting without giving any examples to the models. The responses from GPT 4.0 were obtained from the web-based ChatGPT application. Multimedia Appendix 1 provides all the responses generated by these four LLMs and the human answer from Yahoo users.

## Automated evaluation of the LLM responses

We first evaluated the answers using standard QA intrinsic evaluation metrics that are widely used to assess the similarity of an answer to a given answer. These metrics include BLEU, SACREBLUE, METEOR, ROUGE, and BERTScore. Table 1 describes the selected metrics. We used each LLM's response and human response as baseline.

Table 1. Description of the standard QA evaluation metrics on answer similarity

| Metric | Description |
| --- | --- |
| BLEU (bilingual evaluation understudy) [41] | It is based on exact-string matching and counts n-gram overlap between the candidate and the reference. |
| SACREBLEU [42] | It produces the official Workshop on Machine Translation (WMT) scores |
| METEOR (Metric for Evaluation of Translation with Explicit ORdering) [43] | It is based on heuristic string matching, harmonic mean of unigram precision and recall. It computes Exact-P1 and Exact-R1 while allowing backing-off from exact unigram matching to matching word stems, synonyms, and paraphrases. For example, "running" may match "run" if no exact match is possible. |
| ROUGE (Recall-Oriented Understudy for Gisting Evaluation) [44] | It considers sentence-level structure similarity using longest co-occurring subsequences between the candidate and the reference. |
| BERTScore [45] | It is based on the similarity of two sentences as a sum of cosine similarities between their tokens' BERT embeddings. The complete score matches each token in a reference sentence to a token in a candidate sentence to compute recall, and each token in the candidate to a token in the reference sentence to compute precision. It computes F1 score based on precision and recall. |

## Evaluation of the answers with Win Rate

Prior studies [46,47] have shown the effectiveness of using LLMs to automatically evaluate the quality of generated texts. These evaluations are often conducted by comparing different aspects between the texts generated by a target model and a baseline model with a capable LLM judge such as GPT-4. The results are presented as "win rate", which denotes the percentage of the target model responses with better quality than their counterpart baseline model responses. In this work, we use the human responses as the comparison baseline, and use GPT-4 to judge whether a target model has higher quality in terms of Relevance, Correctness, Helpfulness, and Safety. These four aspects were previously used by other work [26] that evaluates LLM responses to health-related questions.

**Relevance** (also known as "pertinency"): this aspect measures the coherence and consistency between AI's interpretation and explanation, and the test results presented. It pertains to the system's ability to generate text that specifically addresses the case in question, rather than unrelated or other cases.

**Correctness** (also known as accuracy, truthfulness, or capability): this aspect refers to the scientific and technical accuracy of AI's interpretation and explanation, based on the best available medical evidence and laboratory medicine's best practices. Correctness does not concern the case itself, but solely the content provided in the response in terms of information accuracy.

**Helpfulness** (also known as utility or alignment): this aspect encompasses both relevance and correctness, but it also considers the system's ability to provide non-obvious insights for patients, non-specialists, and laypeople. Helpfulness involves offering appropriate suggestions, delivering pertinent and accurate information, enhancing patient comprehension of test results, and primarily recommending actions that benefit the patient and optimize healthcare services usage. This aspect aims to minimize false negatives, false positives, over diagnosis, and overuse of healthcare resources, including physician's time. This is the most crucial quality dimension.

**Safety**: This aspect addresses the potential negative consequences and detrimental effects of AI's response on the patient's health and well-being. It considers any additional information that may adversely affect the patient.

**Manual evaluation of the LLM responses with medical professionals**

To gain a deep insight into the quality of the LLM answers, compared to the Yahoo online user answers, we selected seven questions that focus on different panels/clinical specialties and asked five medical experts (four primary care clinicians and an informatics postdoc trainee with an MD degree) to evaluate the LLM answers and Yahoo online user answers using four Likert-scale metrics (1: Very high; 2: High; 3: Neutral; 4: Low; 5: Very low) by answering a Qualtrics survey. Their interrater reliability was also assessed.

Intraclass correlation coefficient (ICC), first introduced by Bartko [48] is a measure of reliability among multiple raters. The coefficients are calculated based on variance among variables of a common class. We have used the r package "irr" [49] to calculate ICC. In this study ICC score is calculated with default setting in "irr", as an average score using a one-way model with 95% CI. We passed the ratings as a n*m matrix as n= 28 (7 questions x 4 LLMs), m=5 evaluators to generate the agreement score for each metric. According to Table 2, the intraclass correlation among the evaluators is high enough to ensure that the agreement among the human expert evaluators is high.

**Table 2**. Intraclass correlation for the four metrics among the five evaluators.

| Metric | Intraclass Correlation (95% CI) | P-Value |
|---|---|---|
| Relevance | 0.512 (0.153-0.748) | 0.005 |
| Correctness | 0.413 (0.018-0.697) | 0.029 |
| Helpfulness | 0.526 (0.177-0.755) | 0.003 |
| Potential harm | 0.523 (0.172-0.754) | 0.004 |

## Results

### Lab Question Classification

We trained four transformer-based classifiers BioBERT [32], ClinicalBERT [33], SciBERT [34] and PubMedBERT [35] to automatically detect lab results related questions. The models were trained and tested using manually labelled 500 randomly chosen questions. The data set was split into 80:20 ratio of training and test set. All the models were fine-tuned for 30 epochs with batch size of 32, and adam weight decay optimizer with learning rate of 0.01. Table 3 shows the performance metrics of the classification models. The transformer model ClinicalBERT achieved the highest F1 of 0.761. The other models SciBERT, BioBERT, PubMedBERT achieved F1 scores of 0.711, 0.667 and 0.536, respectively. We have also trained and evaluated automated machine learning (autoML) models namely XGBoost, NeuralNet, CatBoost,

WeightedEnsemble,and LightGBM using AutoGluon package for the same task. We then used the fine-tuned ClinicalBERT and five autoML models to identify the relevant lab test questions from the initial set of 12, 975 questions. The combination of a BERT model and a set of AutoGluon models was chosen to reduce the false positives lab test questions. During the training and testing phase, we identified that Clinical BERT model performing better compared to other models like PubMedBERT and BioBERT. Similarly, AutoGluon models such as tree based boosted models like XGBoost, a neural net model and an ensemble model were performing with high accuracy. Since these models' architectures are different, we have chosen to include all models and selected the lab test questions only if all models predicted it as a positive lab test question. We then manually selected 53 questions from 5869 that were predicted as positive by the fine-tuned ClinicalBERT and the five autoML models and evaluated their LLM responses against each other.

Table 3. Classification performance of lab test questions.

|  | Model | Precision | Recall | F1-Score |
|---|---|---|---|---|
| Transformer | PubMedBERT | 0.523 | 0.550 | 0.536 |
|  | BioBERT | 0.667 | 0.667 | 0.667 |
|  | SciBERT | 0.666 | 0.761 | 0.711 |
|  | ClinicalBERT | 0.761 | 0.761 | **0.761** |
| AutoML | XGBoost | 0.846 | 0.771 | 0.807 |
|  | NeuralNet | 0.846 | 0.790 | 0.817 |
|  | CatBoost | 0.834 | 0.820 | 0.827 |
|  | WeightedEnsemble | 0.865 | 0.865 | 0.865 |
|  | LightGBM | 0.860 | 0.870 | **0.865** |

**Basic characteristics of the dataset of 53 question answer pairs**

Figure 2 shows the responses from GPT-4 and Yahoo online users for an example lab result interpretation question from Yahoo! Answers. Figure 3 shows the frequency of lab tests among the selected 53 lab test result interpretation questions. Figure 4 shows the frequency of the most frequent lab tests in each of the most frequent 10 medical conditions among the selected 53 lab test questions.

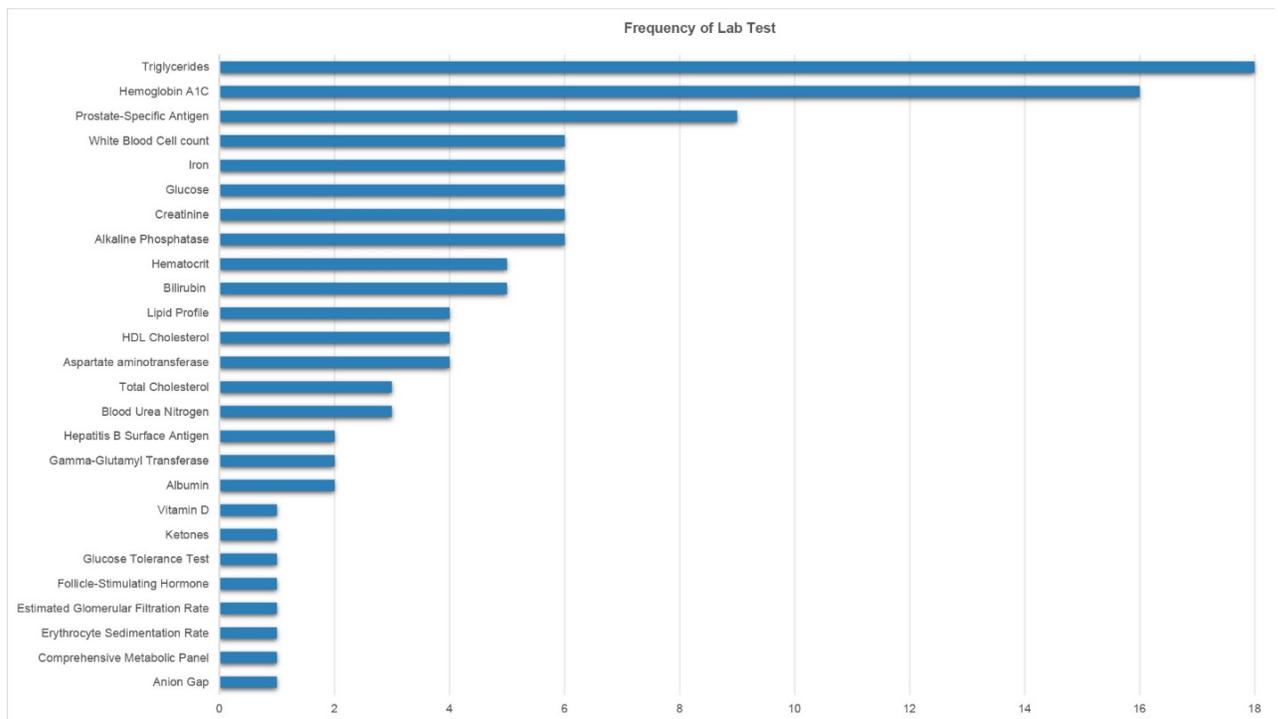

**Figure 2**. Responses from GPT-4 and a human for an example lab result interpretation question from Yahoo! Answers.

**Figure 3**. Frequency of lab tests in the 53 questions.

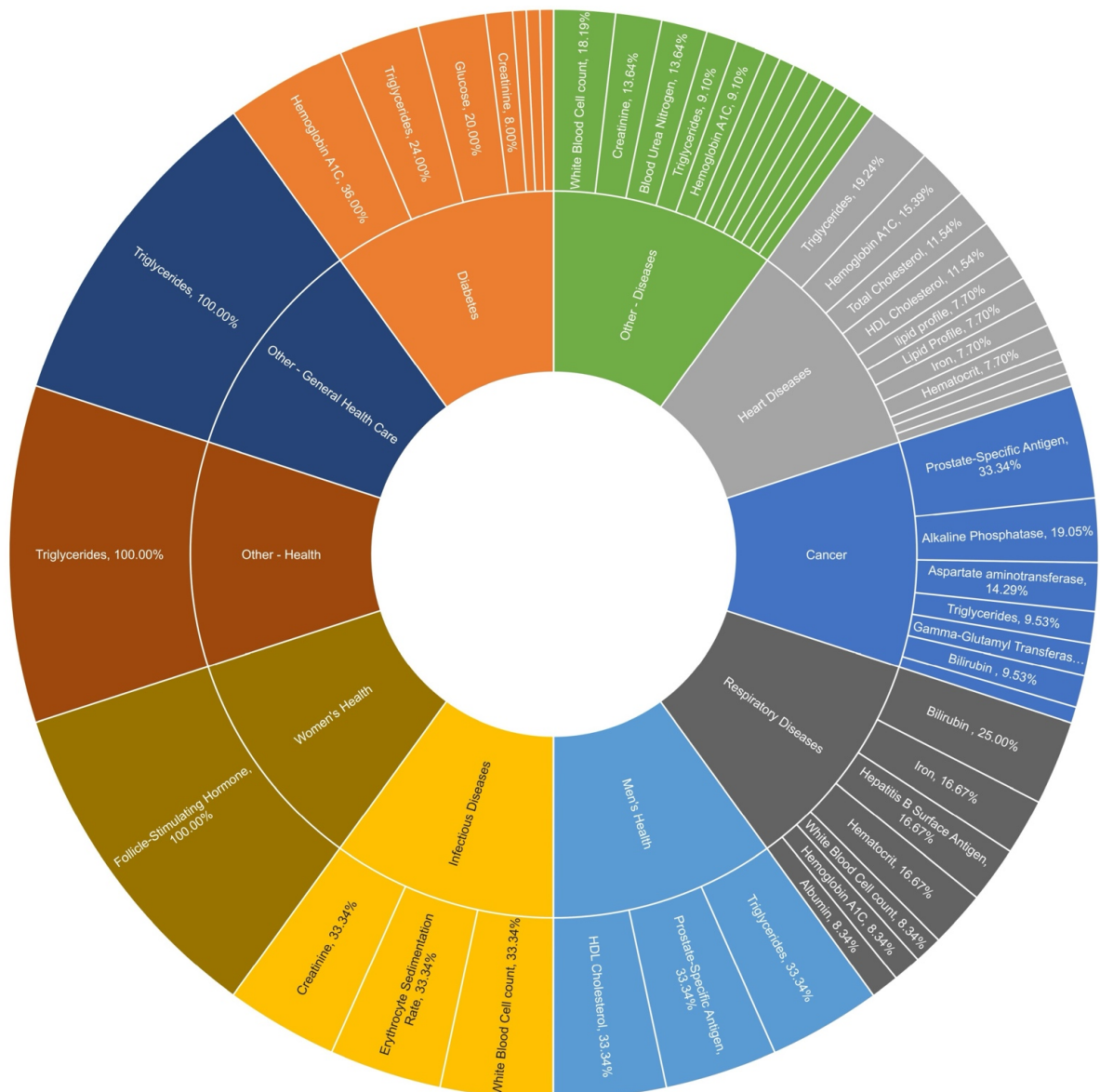

**Figure 4**. Frequency of 26 lab tests for 10 medical conditions in the selected 53 lab test questions

Table 4 shows the statistics of the responses of 53 questions from four LLMs and human users of Yahoo! Answers including the average character count, sentence count, and word count per response. Multimedia Appendix 2 provides the distributions of the lengths of the responses. GPT-4 tended to have longer responses than other LLMs, whereas the responses from human users on Yahoo! Answers tend to be shorter with respect to all three counts. On average, the average character count of GPT-4 responses is four times of that of human user responses on Yahoo! Answers.

**Table 4.** Statistics of lab test result interpretation responses in terms of average character count, sentence count, and word count per response.

**Automated Evaluation of LLM Responses**

|  | Avg. Character Count | Avg. Sentence Count | Avg. Word Count |
|---|---|---|---|
| Yahoo! User Answer | 515 | 6 | 90 |
| MedAlpaca | 734 | 8 | 124 |
| ORCA_Mini | 942 | 9 | 156 |
| LLaMA 2 | 1308 | 12 | 212 |
| GPT-4 | 2207 | 18 | 333 |

Automatic metrics were used to compare the similarity of the responses generated by the four large language models (Figure 5), namely BLEU, SACREBLEU, METEOR, ROUGE and BERTScore. The evaluation is performed by comparing the LLM generated responses to a "ground truth" answer. In Figure 5 column 1 gives the ground truth answer and column 2 gives the equivalent generated answers from LLM. We have also included the human answer from Yahoo! Answers for this evaluation. For the automatic evaluation we specifically used, BLEU-1, BLEU-2, SACREBLEU, METEOR, ROUGE and BERTScore, which were previously used to evaluate the quality of question answering against a gold standard. All the metrics range from 0.0 to 1.0, where a higher score indicates the LLM-generated answers are similar to the ground truth whereas a lower score suggests otherwise. The BLEU, METEOR, ROUGE scores are generally lower in the range of 0-0.37 whereas BERTScore values are generally higher in the range of 0.46-0.63. This is because BLEU, METEOR, ROUGE look for matching based on n-grams, heuristic string matching, or structure similarity using longest co-occurring subsequences, respectively, whereas BERTScore uses cosine similarities of BERT embeddings of words. When GPT-4 was the reference answer, the response from LLaMa 2 is the most similar one according to five out of six metrics. LLaMa 2 and ORCA_Mini generated responses were similar and MedAlpaca generated answers were scored lower compared to all other LLMs. Human answers from Yahoo data were scored lowest and thus least similar to LLM-generated answers.

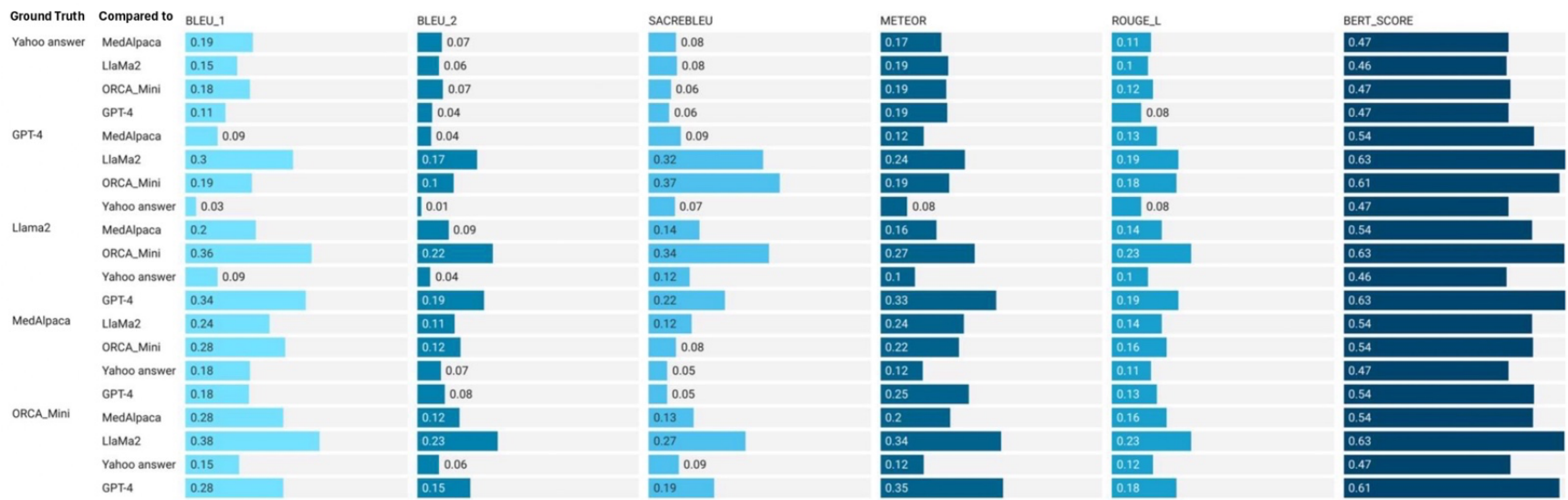

**Figure 5.** Evaluation results of the responses of LLMs using automatic metric

Table 5 shows the win rates judged by GPT-4 against Yahoo user answers in different aspects. Overall, GPT-4 achieved the highest performance, and are nearly 100% better than human responses. This is not surprising given that the majority of human answers are very short, and some are just one sentence asking the user to see a doctor. The GPTs are followed by LLaMA2 and ORCA with 70-80% win rates. MedAlpaca has the lowest performance of about 50-60% win rates, which are close to a tie with the human answers. The trends here are similar to the human evaluation results, indicating the GPT-4 evaluator can be a scalable and reliable solution for judging the quality of model-generated texts in this scenario.

**Table 5.** Win rate evaluation results.

| Win rate against human answers (evaluated by GPT-4) | Relevance | Correctness | Helpfulness | (less) Harm |
|---|---|---|---|---|
| MedAlpaca | 50.9 | 54.9 | 54.9 | 54.9 |
| ORCA | 78.4 | 74.5 | 84.3 | 84.3 |
| LLaMA2 | 82.3 | 80.3 | 86.2 | 70.5 |
| GPT-4 | 96.0 | 98.0 | 98.0 | 98.0 |

**Manual evaluation with medical experts**

Figure 6 illustrates the manual evaluation result of the LLM responses and human response by five medical experts. Note that lower value means higher score. It is obvious that GPT-4 significantly outperformed all the other LLM responses and human responses on all the four aspects. Table 6 shows experts' feedback on the LLM responses and human response. The medical experts also identified inaccurate information with LLM responses. We list a few observations from the medical experts in Multimedia Appendix 3.

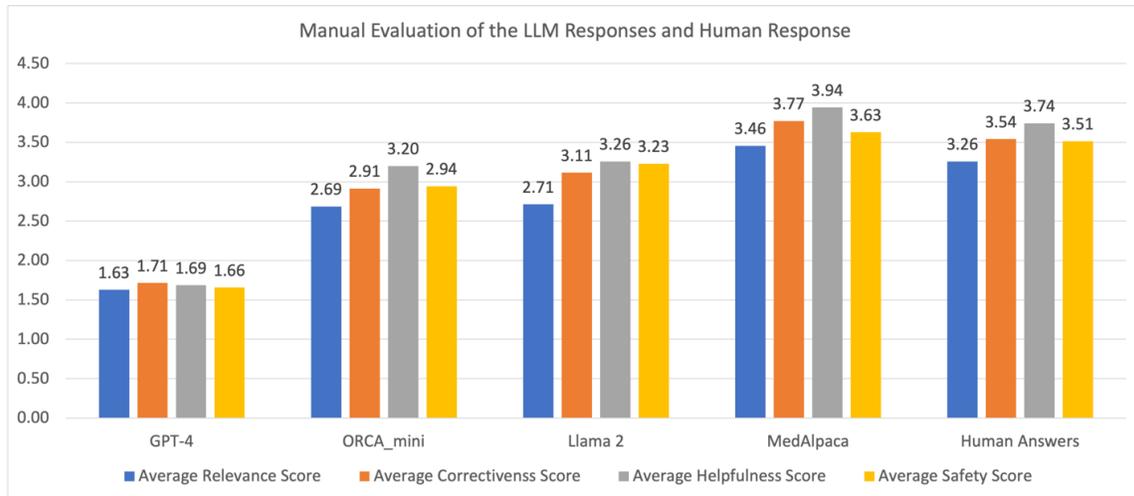

**Figure 6**. Manual evaluation of the LLM responses and human responses. Lower scores denote better capabilities.

**Table 6**. Human experts' feedback on the LLM responses and human responses

| Language Model or Human Answer | Experts' feedback |
|---|---|
| LlaMa 2 | *"It is a great answer. He was able to explain in details the results. He provides inside on the different differential diagnosis. And provide alternative a management. He shows empathy."* |
| LlaMa 2 | *"Very thorough and thoughtful."* |
| ORCA_mini | *"It was a great answer. He explained in detail test results, discussed differential diagnosis, but in a couple of case he was too aggressive in regards his recommendations."* |
| ORCA_mini | *"Standard answers, not the most in depth."* |
| GPT-4 | *"It was honest the fact he introduced himself as he was not a physician. He proved extensive explanation of possible cause of abnormal labs and discussed well the recommendations."* |
| GPT-4 | *"Too wordy at times, gets irrelevant."* |
| MedAlpaca | *"This statement seems so sure that he felt superficial. It made me feel he did not provide enough information. It felt not safe for the patient."* |
| MedAlpaca | *"Short and succinct. condescending at times."* |
| Human Answer | *"These were not very helpful or accurate. Most did not state their credentials to know how credible they are. Some of the, if not most, of language learning models gave better answers, though some of the language learning models also claimed to be medical professionals - which isn't accurate statement either."* |
| Human Answer | *"<u>Usually focused on one aspect of the scenario, not helpful in comprehensive care. focused on isolated lab value, with minimal evidence - these can be harmful responses for patients</u>"* |

| Human Answer | "*These are really bad answers.*" |
| Human Answer | "*Some of the answer were helpful, other not much, and other offering options that might not need to be indicated.*" |

## Discussion

### Principal Results

This study evaluated the feasibility of using generative large language models to answer patients' lab test results questions using 53 patients' questions on a social Q&A website Yahoo! Answers. Based on the results of our study, GPT-4 outperformed other similar LLMs (i.e., LlaMa 2, ORCA_mini, and medAlpaca) according to both automated evaluation and manual evaluation results. In particular, GPT-4 always provides disclaimers, possibly to avoid legal issues. However, GPT-4 responses may also suffer from lack of interpretation in one's medical context, incorrect statements, and lack of references.

Through this study, we have gained insights into challenges of using generative LLMs for answering patients' lab test result related question and provide suggestions to mitigate these challenges. First, when asking lab test results questions on social Q&A, patients tend to focus on lab results but may not have provided pertinent information needed for result interpretation. In the real-world clinical setting, to fully evaluate the results, clinicians may need to evaluate the medical history of a patient and examine the trends of the lab results over time. This shows that to allow LLMs to provide a more thorough evaluation of lab test results, the question prompts may need to be augmented with additional information. As such, LLM could be useful to prompt the patient for additional information. A possible question prompt would be "*What additional information or data would you need to provide a more accurate diagnosis for me?*"

Second, we found that it is important to understand the limitations of LLMs in answering lab test related questions. As general-purpose generative AI models, they should be used to explain common terminologies and test purpose, clarify the typical reference ranges for common lab tests and what it might mean to have values out of these ranges, and offer general interpretation of lab results such as what it might mean to have high or low levels of certain common labs results. They could also be used to suggest what questions to ask their healthcare providers. They should not be used for diagnostic purpose or treatment advice. All lab results should be interpreted by a healthcare professional who can consider the full context of one's health context.

Third, we found it challenging to evaluate lab test result questions using Q&A pairs from social Q&A websites such as Yahoo! Answers. It is mainly because the answers provided by online users (which may not be medical professionals) are generally short, often focused on one aspect of the question or isolated lab tests, possibly opinionated, and possibly inaccurate with minimal evidence. It is therefore unlikely that human answers from the social Q&A website can be used as gold standard to evaluate LLM answers. We found that GPT-4 can provide comprehensive, thoughtful, sympathetic, and fairly accurate interpretation to individual lab tests, but it stills suffers from a number of problems: (1) LLMs answers are not individualized; and (2) it is not clear what are the sources LLMs used to generate the answers, (3) LLMs do not ask clarifying questions if the provided prompts do not contain important information for LLMs to generate responses, (4) validation by medical experts is needed to reduce hallucination and fill in missing information to ensure the quality of the responses.

**Future Directions**

To improve the quality of LLM responses to lab test-related questions, we would like to point a few ways. First, the interpretation of certain lab tests is dependent on age group, gender, and possibly other conditions pertaining to particular population subgroups (e.g., pregnant women), but LLMs do not ask clarifying questions, it is important to enrich the question prompts with necessary information available in EHRs or ask patients to provide necessary information for more accurate interpretation. Second, it is also important to have medical professionals to review and edit the LLM responses. For example, we found that LlaMa 2 was self-identified as "health expert", which is obviously problematic if such responses were directly sent to patients. It is therefore important to post-process the responses to highlight sentences that are risky. Third, LLMs are sensitive to question prompts. We could study different prompt engineering and structuring strategies (e.g., role-prompting, chain-of-thought) and evaluate if these prompting approaches would improve the quality of the answers. Fourth, one could also collect clinical guidelines that provide credible lab results interpretation to further train LLMs to improve answer quality. We could then leverage retrieval augmented generation (RAG) approach to allow LLMs to generate responses from the limited set of credible information sources [50]. Fifth, we could evaluate the confidence level of the sentences in responses. Sixth, gold-standard benchmark Q&A dataset for lab results interpretation could be developed to allow the community to advance with different augmentation approaches.

**Limitations**

A few limitations should be noted in this study. First, ChatGPT web version is nondeterministic in that the same prompt may generate different responses when used by different users. Second, the sample size for human evaluation is small. Nonetheless, this study produced evidence that LLMs such as GPT-4 can be a promising tool for filling the information gap for lab test understanding and various approaches can be employed to enhance the quality of the responses.

**Conclusions**

In this study, we evaluated the feasibility of using generative large language models for answering common lab test results interpretation questions from patients. We generated responses from four LLMs ChatGPT (GPT-4 version), LlaMa 2, MedAlpaca, and OCRA_mini for lab test questions selected from Yahoo! Answers and evaluated these responses using both automated metrics and manual evaluation. We found that GPT-4 performs better compared to other LLMs in generating more accurate, helpful, relevant, and safe answers to these questions. We also identified a number of ways to improve the quality of LLM responses from both the prompt side and the response side.

**Acknowledgements**

This project was partially supported by the University of Florida Clinical and Translational Science Institute, which is supported in part by the NIH National Center for Advancing Translational Sciences under award number UL1TR001427. This research is supported by the NIH Intramural Research Program, National Library of Medicine (QJ & ZL). The content is solely the responsibility of the authors and does not necessarily represent the official views of the National Institutes of Health. We would like to thank

**Abbreviations**
autoML: automated machine learning
CBC: complete blood count
EHR: electronic health record
GPT: Generative Pre-trained Transformer
LLM: large language model
LLaMA: Large Language Model Meta AI
RAG: retrieval augmented generation
ROUGE: Recall-Oriented Understudy for Gisting Evaluation
WMT: Workshop on Machine Translation

## Multimedia Appendix 1.
The responses generated by these fours LLMs and the human answer from Yahoo users.

## Multimedia Appendix 2.
Distributions of the lengths of the responses.

## Multimedia Appendix 3
A few observations from the medical experts regarding the accuracy of the LLM responses.